\documentclass[10pt,twocolumn,letterpaper]{article}

\usepackage[pagenumbers]{iccv} 

\usepackage{xcolor}
\usepackage{microtype}
\usepackage{xcolor,soul}
\usepackage{bbm}
\usepackage[hang,flushmargin]{footmisc}
\usepackage{tikz}
\usetikzlibrary{calc, positioning, spy}
\newlength{\figurewidth}

\usepackage{threeparttable}
\usepackage{tablefootnote} 
\usepackage[accsupp]{axessibility}
\definecolor{iccvblue}{rgb}{0.21,0.49,0.74}
\usepackage[pagebackref,breaklinks,colorlinks,allcolors=iccvblue]{hyperref}

\def\paperID{8828} 
\def\confName{ICCV}
\def\confYear{2025}

\title{DeGauss: Dynamic-Static Decomposition with Gaussian Splatting for Distractor-free 3D Reconstruction}
\author{Rui Wang \quad Quentin Lohmeyer\quad Mirko Meboldt\quad Siyu Tang\vspace{0.3em} \\
{\normalsize ETH Z\"urich} \\
{\tt\small \{ruiwang46, qlohmeye, meboldt\}@ethz.ch} \quad\quad {\tt\small siyu.tang@inf.ethz.ch} \quad\quad\quad\quad\\
{\normalsize \url{https://batfacewayne.github.io/DeGauss.io/}}
\vspace{-5mm}
}
\begin{document}
\definecolor{commentblue}{rgb}{0.0196,    0.4863,    0.8784}
\newcommand{\rui}[1]{{\color{commentblue}[rui: {#1}]}}

\definecolor{lightblue}{rgb}{.8,.95,1}
\newcommand{\hb}[1]{\hl{\textbf{#1}}}
\newcommand{\lpips}{\scalebox{0.8}{LPIPS$\downarrow$}}
\newcommand{\ssim}{\scalebox{0.8}{SSIM$\uparrow$}}
\newcommand{\psnr}{\scalebox{0.8}{PSNR$\uparrow$}}
\newcommand{\first}[1]{{\colorbox{red!40}{#1}}}
\newcommand{\second}[1]{{\colorbox{orange!50}{#1}}}
\newcommand{\third}[1]{{\colorbox{yellow!50}{#1}}}
\twocolumn[{
\renewcommand\twocolumn[1][]{#1}%
\maketitle
  \begin{center}
  \centering
    \includegraphics[width=\textwidth]{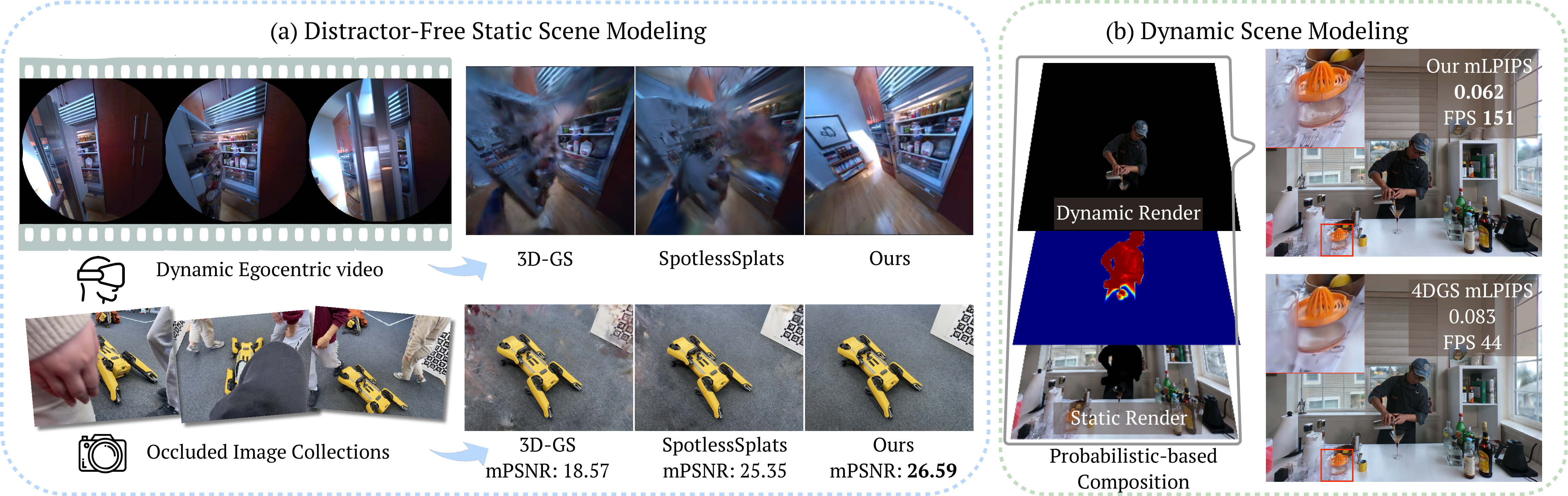}
    \captionof{figure}{With self-supervised foreground-background gaussian splats modeling and accurate decomposition, \textbf{DeGauss} simultaneously enables \textbf{(a)}: SOTA distractor-free static scene reconstruction for casual captures (no dynamic modeling in the static background) and \textbf{(b)}: efficient, high-quality dynamic-static representation for dynamic scenes (no static modeling in dynamic foreground).}
    \label{fig:teaser}
  \end{center}
}]
\begin{abstract}
Reconstructing clean, distractor-free 3D scenes from real-world captures remains a significant challenge, particularly in highly dynamic and cluttered settings such as egocentric videos. To tackle this problem, we introduce DeGauss, a simple and robust self-supervised framework for dynamic scene reconstruction based on a decoupled dynamic-static Gaussian Splatting design. DeGauss models dynamic elements with foreground Gaussians and static content with background Gaussians, using a probabilistic mask to coordinate their composition and enable independent yet complementary optimization. DeGauss generalizes robustly across a wide range of real-world scenarios, from casual image collections to long, dynamic egocentric videos, without relying on complex heuristics or extensive supervision. Experiments on benchmarks including NeRF-on-the-go, ADT, AEA, Hot3D, and EPIC-Fields demonstrate that DeGauss consistently outperforms existing methods, establishing a strong baseline for generalizable, distractor-free 3D reconstruction in highly dynamic, interaction-rich environments.
\end{abstract}    
\section{Introduction}
\label{sec:intro}
Recent advances in Neural Radiance Fields (NeRF)~\cite{mildenhall2021nerf} and 3D Gaussian Splatting~\cite{kerbl3Dgaussians} have enabled scalable 3D scene reconstruction and high-quality novel view synthesis from image collections. However, these methods perform well primarily on datasets captured under controlled conditions, where scenes remain mostly static and consistent across views. They struggle to generalize to casual captures containing dynamic elements, such as moving objects and humans. In such cases, dynamic content is often modeled as view-dependent artifacts, resulting in numerous "floaters" in the reconstructed scene.

This limitation is further amplified in egocentric videos, a rapidly growing data source that introduces unique challenges for 3D scene reconstruction\cite{lv2024aria, sun2023aria,tschernezki2024epic,gu2025egolifter,zhang2024egogaussian}. Egocentric videos, typically recorded with head-mounted, forward-facing cameras, are characterized by rapid, embodied motion. Besides substantial camera movement and motion blur, these videos frequently capture dynamic objects that the camera wearer interacts with, as well as the wearer's own body. These factors introduce significant challenges for standard scene reconstruction methods.

The key question we aim to address in this work is how to reconstruct clean, distractor-free 3D scenes from real-world, in-the-wild videos. 
We focus on developing a robust and generalizable framework capable of handling a wide range of everyday capture scenarios, from casual, uncontrolled image collections to long-duration, highly dynamic egocentric recordings. By explicitly tackling the presence of dynamic elements, we aim to push 3D scene reconstruction beyond static environments toward realistic, interaction-rich settings. 

To model dynamics in 3D reconstruction, recent methods such as NeRF-on-the-go, WildGaussians, and SpotlessSplats~\cite{ren2024nerf,kulhanek2024wildgaussians,sabour2024spotlesssplats} propose to suppress transient regions during training, achieving state-of-the-art distractor-free scene reconstruction on casual image collections. These approaches leverage reconstruction loss residuals and semantic features~\cite{oquab2023dinov2,tang2023emergent} to identify and mask dynamic content, as transient regions often exhibit higher reconstruction errors. However, these methods typically rely on careful initialization and stable optimization, which limits their ability to handle the complex dynamics of egocentric videos, where continuous human-scene interactions, severe motion blur, and rapid illumination changes make static-dynamic separation particularly challenging.

Meanwhile, several self-supervised NeRF-based methods aim to jointly model dynamic and static elements through explicit dynamic branches and masking strategies~\cite{martin2021nerf,tschernezki2021neuraldiff,wu2022d}. While these methods improve generalization across diverse inputs, they suffer from long training times and struggle to balance dynamic and static representations. 
For 3D scenes captured with temporally sparse image inputs, the dynamic branch may fail to fully segment dynamic elements, leaving floaters in the static reconstruction~\cite{sabour2023robustnerf}. In contrast, for highly dynamic egocentric videos, the dynamic branch often over-segments dynamic regions, dominating the reconstruction and leaving the static scene under-represented~\cite{pan2023aria}. 

In this work, we propose \textbf{DeGauss}: Dynamic-Static Decomposition with Gaussian Splatting for Distractor-free 3D Reconstruction. It is a simple and robust self-supervised framework that leverages dynamic-static Gaussian Splatting to effectively model and separate dynamic elements from input scenes. \textbf{DeGauss} generalizes across a wide range of scenarios, from casual image collections such as the NeRF-on-the-go dataset~\cite{ren2024nerf} to highly dynamic egocentric video sequences like ADT~\cite{pan2023aria}, AEA~\cite{lv2024aria}, Hot3D~\cite{pan2023aria}, and EPIC-Fields~\cite{tschernezki2024epic}, consistently delivering superior performance without complex heuristics or elaborate designs.

Our key insight is to leverage the complementary strengths of dynamic and static Gaussians for coordinated optimization for dynamic scene reconstruction. 
Specifically, dynamic Gaussian methods~\cite{yang2023deformable3dgs,wu20244d} learn deformation fields for temporal modeling but tend to overfit to training views and generalize poorly to novel viewpoints~\cite{stearns2024dynamic,gao2022monocular}. In contrast, static Gaussians, while limited in handling motion, offer more stable representations across views, modeling dynamic elements as view-dependent artifacts (e.g., floaters).
To combine their advantages, we propose a decoupled foreground-background Gaussian representation, where dynamic elements are modeled with foreground Gaussians and static content with background Gaussians. A probabilistic mask, rasterized from the foreground Gaussians, controls the composition of the two branches and enables coordinated yet independent optimization. During training, ambiguous regions are updated jointly, while floaters in the static branch are progressively suppressed through partial opacity resets and pruning. To further improve robustness under varying illumination, we introduce a brightness control mask to enhance non-Lambertian effects modeling capability of the background branch during training and mitigate dynamic-static ambiguities in those regions. Beyond producing clean, distractor-free 3D reconstructions, our formulation offers an efficient, hybrid representation of dynamic scenes through this decoupled dynamic-static design. We show that our method achieves superior results compared to baseline dynamic scene modeling approaches, with notable advantages across diverse datasets~\cite{li2022neural,park2021hypernerf}.
In summary, our contributions are: 
\begin{itemize}
    \item We propose \textbf{DeGauss}, a decoupled foreground-background design which leverages dynamic-static Gaussian splatting for robust and generalizable dynamic-static decomposition.
    \item Our proposed method achieves state of the art distractor-free reconstruction results for both highly challenging egocentric videos and image collections.
    \item We demonstrate that \textbf{DeGauss} enables high-quality and efficient dynamic scene modeling through the decoupled dynamic-static representation.
\end{itemize}

\section{Related Work}
\label{sec:related}
\textbf{Distractor-Free Neural Reconstruction} based on loss residual of input images and renders during reconstruction was investigated in~\cite{sabour2023robustnerf,chen2024nerfhugs}. In~\cite{gu2025egolifter}, it is additionally combined with open-world 3D segmentation task with Segment Anything masks~\cite{kirillov2023segment}.
NeRF-on-the-Go~\cite{ren2024nerf} leverages DINOV2 features~\cite{oquab2023dinov2}, color residuals, and an MLP predictor for dynamic elements mask. This approach was later extended to gaussian splatting~\cite{kerbl3Dgaussians} in WildGaussians~\cite{kulhanek2024wildgaussians}. SpotlessSplats~\cite{sabour2024spotlesssplats} utilizes clustered diffusion-based features~\cite{tang2023emergent} and SOTA distractor-free scene modeling for image collections. However, these methods are sensitive to initialization and fail to leverage semantic information when within-class dynamic-static ambiguities or scene deformations arise, which limits their generalizability in more challenging settings.
\\
\textbf{Self-Supervised Scene Decomposition} for neural fields was first introduced in Nerf-W~\cite{martin2021nerf}, which decomposes and models the whole scene with dynamic and static neural fields. This approach is further generalized to egocentric videos in NeuralDiff~\cite{tschernezki2021neuraldiff}, decomposing the entire scene into dynamic, static, and actor branches. D$^2$NeRF~\cite{wu2022d} enhances decomposition results for small scenes and short clips by incorporating assignment regularization and a shadow field. However, in general, these methods face balancing issues between static-dynamic reconstruction and do not generalize well to long video inputs.\\
\textbf{Dynamic Gaussian Splatting} modeling via explicit trajectory modeling to track gaussian dynamic was investigated in~\cite{luiten2023dynamic,huang2023sc}. Deformable-GS~\cite{yang2023deformable3dgs} employs a deformation network to encode Gaussian deformations. 4DGS~\cite{wu20244d} leverages a Hex-plane\cite{cao2023hexplane} encoder and MLP-based decoders to model time-dependent Gaussian attribute parameters. However, these methods struggle to predict different deformations for gaussians with proximity, leading to over-smoothed dynamic motion.
A Recent method~\cite{wuswift4d} tackles this with dynamic-static separation by pre-computing static-dynamic decomposition masks
based on video pixel intensity variation. However, this method only works for fixed-view camera inputs with simple motion.\\
\textbf{Concurrent work}: Recent methods~\cite{wang2024desplat,lin2025hybridgs} fit separate per-camera-space gaussians for every training view to model and segment out dynamic elements with self-supervised modeling for image collections. However, the lack of shared distractors modeling across images makes it sensitive to initialization and hard to generalize. With foreground dynamic gaussians, our method achieves SOTA distractor-free scene reconstruction results for both challenging egocentric videos~\cite{lv2024aria,banerjee2024introducing,pan2023aria,tschernezki2024epic} and casual image collections~\cite{ren2024nerf}.

\section{Method}
\label{sec:method}

\begin{figure*}[t]
	\centering
	\includegraphics[width=1\linewidth]{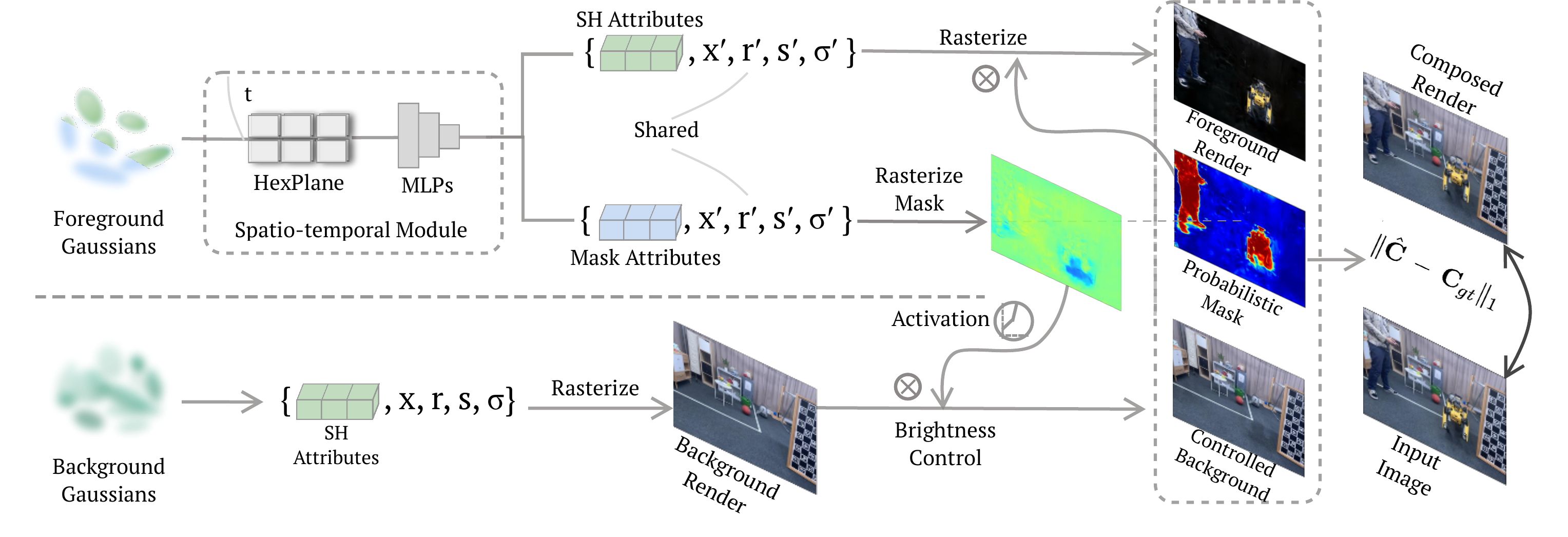}
    \vspace{-4ex}
	\caption{Our method simultaneously reconstructs the 3D scene and learns an unsupervised decomposition into decoupled static background and dynamic foreground branches, where the update is loosely controlled by the mask rasterization process. This decoupled formulation guarantee flexible yet accurate scene decomposition result.}
	\label{fig:pipeline}
\vspace{-1ex}
\end{figure*}

\subsection{3D Gaussian Splatting}
\label{subsec:3D Gaussian Splatting}
3D Gaussian Splatting~\cite{kerbl3Dgaussians} provides an explicit representation of a 3D scene using Gaussian primitives. Each primitive is defined by a mean vector \(\mathbf{x} \in \mathbb{R}^3\) and a covariance matrix \(\mathbf{\Sigma} \in \mathbb{R}^{3 \times 3}\), where
\begin{align}
    \label{eqn:gaussian_geometry}
    \mathcal{G} ({x}) &= \exp \Bigl( -\frac{1}{2} \left( {x} - \mathbf{x} \right)^T \Sigma^{-1} \left( {x} - \mathbf{x} \right) \Bigr), \\
    \text{s.t.} \quad \Sigma &= \mathbf{R}\text{diag}(\mathbf{s})\text{diag}(\mathbf{s})^T \mathbf{R}^T \nonumber
\end{align}
with \(\mathbf{R}\) being the are rotation matrix 
that could be represented by quaternion $\mathbf{r}$ 
and \(\mathbf{s}\) being the scale vector.

To render these Gaussians onto the image plane, we use differentiable splatting~\cite{yifan2019differentiable}, which applies a projection transformation \(\mathcal{P}(\mathcal{G})\). The final color \(\mathbf{C}\) at each pixel is then computed by blending the contribution of all Gaussians, sorted by their depth:
\begin{equation}
    \label{eqn:gaussian_splatting}
    \mathbf{C}
    = \sum_{i=1}^{N} \mathbf{c}_i \,\sigma_i \,\mathcal{P}_i(\mathcal{G}_i)
      \prod_{j=1}^{i-1}(1 - \sigma_j \,\mathcal{P}_j(\mathcal{G}_j)).
\end{equation}
Here, \(\mathbf{c}_i\in\mathbb{R}^k\) are spherical harmonic (SH) coefficients (for an SH basis of degree \(k\)), and \(\sigma_i\in\mathbb{R}\) denotes the opacity of the \(i\)th Gaussian.

\subsection{Foreground deformable gaussian}
\label{subsec:4dgs}
We extend the set of foreground Gaussians $\mathcal{G}_f$ to embed customized mask elements for dynamic scene decomposition, and the complete features could be defined as $\mathcal{G}_f=\{\mathbf{x},\mathbf{s}, \mathbf{r},\mathbf{\sigma},\mathbf{c},m_f,m_b,b\}$. Here, the standard attributes $\{m_f, m_b, b\}$ are the foreground probabilistic attributes, background probabilistic attributes, and brightness control attributes, respectively.

The deformed foreground Gaussians are obtained as:
$\mathcal{G}_f^\prime = \Delta\mathcal{G}_f + \mathcal{G}_f$. The spatial-temporal module comprises an encoder $\mathcal{H}$ and a decoder $\mathcal{D}$. The encoder, based on Hexplane~\cite{cao2023hexplane}, extracts spatio-temporal features based on reference time $t$ with $\mathbf{f_d} = \mathcal{H}(\mathcal{G}_f, t)$, and the multi-head decoder $\mathcal{D}$ predicts the deformation of each gaussian features with $\Delta \mathcal{G}_f = \mathcal{D}(\mathbf{f_d})$. Separate MLPs are employed to predict the deformation of each gaussian attribute. The decoder $\mathcal{D}$ comprises:
 $\mathcal{D}=\{\phi_\mathbf{x}, \phi_\mathbf{r}, \phi_\mathbf{s}\, \phi_\sigma, \phi_\mathbf{c}, \phi_{m_f}, \phi_{m_b}, \phi_b \}$. With this, the deformed feature could be addressed as:
\begin{align}
(\mathbf{x}^\prime, \mathbf{r}^\prime,& \mathbf{s}^\prime, \sigma^\prime, \mathbf{c}^\prime, m_f^\prime, m_b^\prime, b^\prime)
= \Bigl( \mathbf{x} + \phi_\mathbf{x}(\mathbf{f}_d),\, \nonumber \\
&\mathbf{r} + \phi_\mathbf{r}(\mathbf{f}_d),\, 
\mathbf{s} + \phi_\mathbf{s}(\mathbf{f}_d), \sigma + \phi_\sigma(\mathbf{f}_d),\, \mathbf{c} + \phi_\mathbf{c}(\mathbf{f}_d),\,\nonumber\\
& m_f + \phi_{m_f}(\mathbf{f}_d), m_b + \phi_{m_b}(\mathbf{f}_d),\, b + \phi_b(\mathbf{f}_d) \Bigr).
\end{align}
\subsection{Probabilistic Composition Mask Rasterization}
Given the predicted mask elements \(\{m_f^\prime, m_b^\prime\}\) and the deformed attributes \(\{\mathbf{x}^\prime, \mathbf{r}^\prime, \mathbf{s}^\prime, \sigma^\prime\}\), we can directly use differentiable rendering to compute the raw foreground probability \(\mathbf{M}_f\)  and $\mathbf{M}_b$ via based on~\cref{eqn:gaussian_splatting}:
\begin{align}
    \mathbf{M}_f = \sum_{i=1}^{N} {m_f^\prime}_i \sigma_i^\prime \mathcal{P}_i(\mathcal{G}_{f_i}^\prime) \prod_{j=1}^{i-1} (1 - \sigma_j^\prime\mathcal{P}_j(\mathcal{G}_{f_j}^\prime)), \\
    \mathbf{M}_b = \sum_{i=1}^{N} {m_b^\prime}_i \sigma_i^\prime \mathcal{P}_i(\mathcal{G}_{f_i}^\prime) \prod_{j=1}^{i-1} (1 - \sigma_j^\prime\mathcal{P}_j(\mathcal{G}_{f_j}^\prime)); 
\end{align}
With $\mathbf{P} =M_f + M_b +\epsilon$, where $\epsilon$ is a small constant to avoid division by zero, the foreground and background probabilistic masks could be given by:
\begin{equation}
    \mathbf{P}_f = (1/\mathbf{P}) * \mathbf{M}_f, P_b = (1/\mathbf{P}) * \mathbf{M}_b.
\end{equation}
This probabilistic formulation naturally discourages mid-range values (near 0.5), pushing the prediction toward 0 or 1 and yielding a clean dynamic-static decomposition.
\subsection{Background Brightness Control}
Casual captures often exhibit significant illumination variations, creating ambiguities in geometry and view-dependent appearance modeling. While non-Lambertian effects can be progressively captured through the spherical harmonic (SH) coefficients of Gaussian Splatting, the high expressiveness of dynamic Gaussians in the foreground branch often leads to over-segmentation of dynamic elements in regions with large illumination variations. To address this, we introduce a brightness control mask that enhances the background branch’s capacity to model non-Lambertian effects. The raw brightness control mask could be obtained via rasterizing foreground gaussian with brightness control element $b$:
\begin{equation}
\mathbf{B} = \sum_{i=1}^{N} {b^\prime}_i \sigma_i^\prime \mathcal{P}_i(\mathcal{G}_{f_i}^\prime) \prod_{j=1}^{i-1} (1 - \sigma_j^\prime\mathcal{P}_j(\mathcal{G}_{f_j}^\prime))
\end{equation}
Moreover, to prevent modeling dark dynamic objects with the brightness control mask and enable the modeling of over-brightness, we further introduce a piece-wise linear activation function for the brightness control mask, and the transformed brightness control mask $\mathbf{\hat{B}}$ is given by:
\begin{align}
\hat{\mathbf{B}}=
\begin{cases}
\mathbf{B}+0.5, &0\le \mathbf{B}\le 0.75,\\
k\,(\mathbf{B}-0.75)+1.25, &0.75< \mathbf{B}\le 1,
\end{cases}
\end{align}
, where k is an over-brightness modeling coefficient, we choose $k=35$ in practice. The raw background render $\mathbf{C}_b$ is rasterized by background gaussian $\mathcal{G}_b$ with equation~(\ref{eqn:gaussian_splatting}). The controlled background is then given with $\mathbf{\hat{C}}_b=\mathbf{\hat{B}} * \mathbf{C}_b$.
\subsection{Dynamic Foreground Representation}
With deformed gaussian $\mathcal{G}_f^\prime$, the raw foreground render could be given by:
\begin{align}
    \mathbf{C}_f(u,v) &= \sum_{i=1}^{N} {\mathbf{c}_f^\prime}_i \sigma_i^\prime \mathcal{P}_i(\mathcal{G}_{f_i}^\prime) \prod_{j=1}^{i-1} (1 - \sigma_j^\prime\mathcal{P}_j(\mathcal{G}_{f_j}^\prime)), 
\end{align}
And the final foreground render $\mathbf{\hat{C}}_f$ is obtained by applying the foreground probabilistic mask to the raw foreground render,  $\mathbf{C}_f = \mathbf{P}_f\mathbf{C}_f$.
This formulation comes with several advantages. On one hand, we could efficiently allow the presence of utility gaussians that are important for probabilistic composition mask $\mathbf{P}_f$, $\mathbf{P}_b$ and brightness control mask $\mathbf{\hat{B}}$ but do not contribute to foreground render. Moreover, such a design could efficiently reduce the presence of unregulated gaussian movement for dynamic scene modeling with this added degree of freedom and avoid artifacts caused by unconstrained gaussian movement.
\subsection{Unsupervised scene decomposition}
\begin{figure}[t]
	\centering
	\includegraphics[width=1\linewidth]{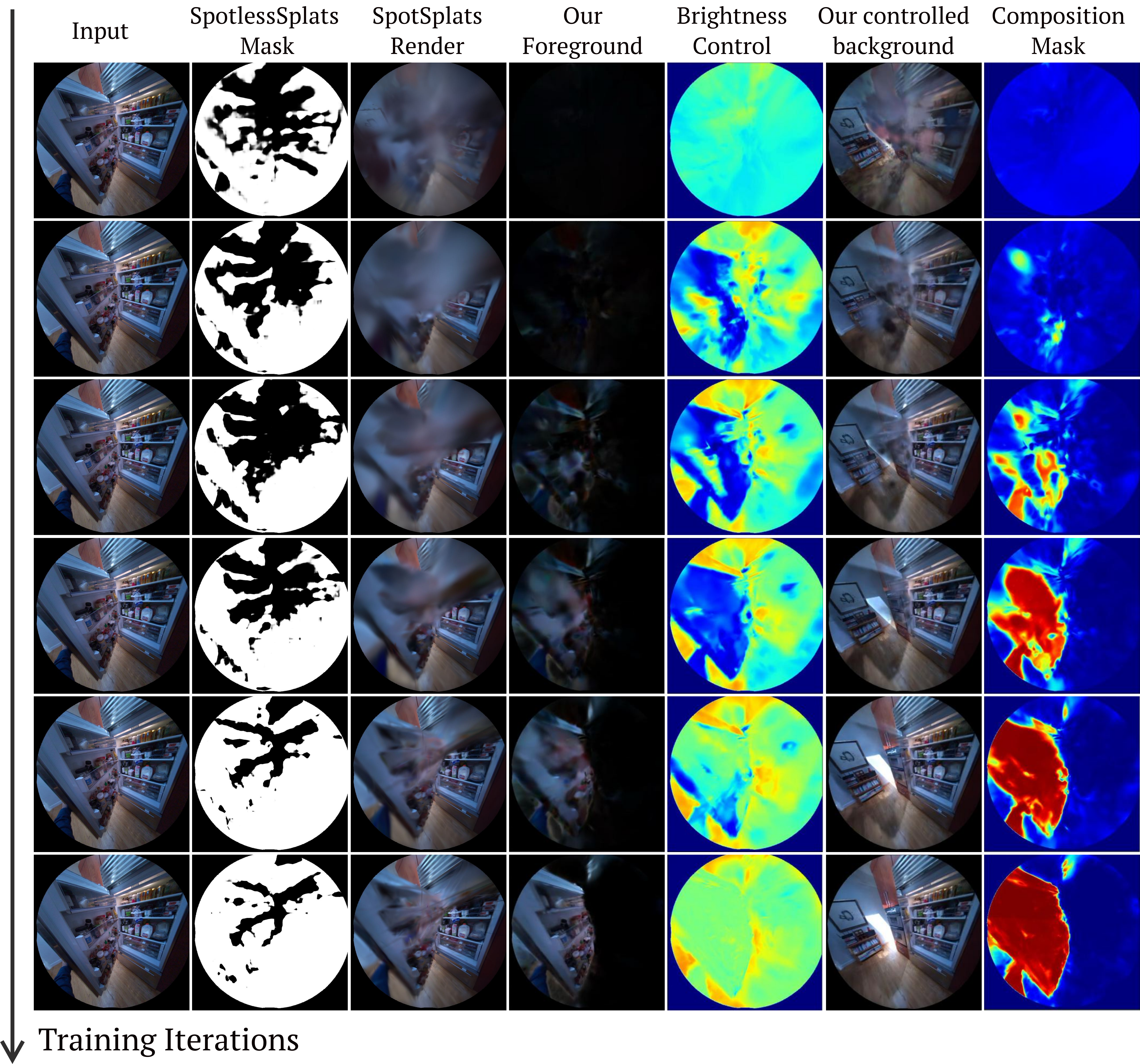}
	\caption{Compared to SpotlessSplats \cite{sabour2024spotlesssplats}, which is constrained by initialization and overfit to floaters. Our method offers significantly greater robustness in handling local minimas. The brightness control mask effectively resolves the static-dynamic ambiguity due to strong illumination variations and promote the decomposition process during training.}
	\label{fig:training}
\end{figure}

With the established composition mask $\mathbf{P}_f$, $\mathbf{P}_b$ and brightness control mask $\hat{\mathbf{B}}$, the composed render $\hat{C}$ is defined as:
\begin{equation}
    \hat{\mathbf{C}}= \mathbf{P}_{f}*\mathbf{C}_f + \mathbf{P}_b*\hat{\mathbf{B}} * \mathbf{C}_b
\end{equation}
Compositional rendering with color mixing in NeRF-based methods~\cite{li2022neural,tschernezki2021neuraldiff,martin2021nerf} sorts and integrates static and dynamic density and radiance along each ray(compose during rendering), leading to early ray termination during training on local minima and reconstructs static scene without fine details\cite{tschernezki2021neuraldiff,sabour2023robustnerf}. In our decoupled design, the dynamic/static gaussians rasterize the foreground/background renders $\mathbf{C}_f$ and $\mathbf{C}_b$ independently and compose (after rending) with the probabilistic mask $\mathbf{P}_f$. This design enables full gradient flow and allow gradually formulated composition mask during training, as shown in~\cref{fig:training}. The wrongly modeled elements are gradually pruned during gaussian density control process, yielding accurate, clean yet flexible dynamic-static separation results that is much more robust to local minima.
\subsection{Loss function}
Loss function design is crucial to balance the expressiveness of foreground and background branches while reconstructing the scene with high-quality details. As the loss gradient magnitude controls the densification process of gaussians~\cite{kerbl3Dgaussians}, we design the loss function $\mathcal{L}$, which comprises two parts $\mathcal{L}_\text{main}$ and $\mathcal{L}_\text{uti}$, separating loss gradients for adaptive densification process, to effectively suppress the spawning of floaters and controlling the number of utility gaussians in foreground branch:
\begin{align}
\mathcal{L} =& \underbrace{\mathcal{L}_{\text{1}}+\mathcal{L}_\text{diversity} + \mathcal{L}_\text{reg} +\mathcal{L}_\text{depth} + \mathcal{L}_f + \mathcal{L}_b}_{\mathcal{L}_\text{main} }\nonumber\\
&+ \underbrace{\mathcal{L}_\text{SSIM} + \mathcal{L}_\text{entropy}+\mathcal{L}_\text{brightness} + \mathcal{L}_\text{scale}}_{\mathcal{L}_\text{uti}}.
\end{align}
While both main loss $\mathcal{L}_\text{main}$ and utility loss $\mathcal{L}_\text{uti}$ are used for optimizable parameters' update, 
only the gradient magnitude of $\mathcal{L}_\text{main}$ are used to densify foreground and background gaussians. We refer readers to Appendix~\textbf{A.} for a detailed definition of each loss term.

\subsection{Partial Opacity Reset}
In methods as~\cite{sabour2024spotlesssplats,kulhanek2024wildgaussians}, directly employing periodic opacity reset~\cite{kerbl3Dgaussians} is not feasible, as it induces instability during training. Owing to the added stability with the foreground-background formulation, we perform periodic partial opacity reset for $50\%$ for background-foreground gaussians. This guarantees stable training, effectively controls gaussian density, and handles local minima. 
\section{Experiments}
\label{sec:experiments}
\input{figs/aria}
\input{figs/qualiNerfonthego}
\begin{table*}[htbp]
    \begin{center}
    \caption{Distractor free scene reconstruction on NeRF On-the-go Dataset\cite{ren2024nerf}.The \first{best}, \second{second best}, and \third{third best} are highlighted. $\ddagger$: $\pm$0.005  SSIM and LPIPS due to rounding uncertainty of originally reported result. Our method shows generally superior performance over state-of-the-art methods.}
    \label{table:onthego}
    \scalebox{0.5}{
    \begin{tabular}{ccccccccccccccccccc|ccc}
    \toprule
    \multicolumn{1}{c}{}  & \multicolumn{3}{c}{Mountain}  & \multicolumn{3}{c}{Fountain} & \multicolumn{3}{c}{Corner}  & \multicolumn{3}{c}{Patio} & \multicolumn{3}{c}{Spot} & \multicolumn{3}{c|}{Patio-High} & \multicolumn{3}{c}{Mean} \\
    \cmidrule(lr){2-4} \cmidrule(lr){5-7} \cmidrule(lr){8-10} \cmidrule(lr){11-13} \cmidrule(lr){14-16} \cmidrule(lr){17-19} \cmidrule(lr){20-22}
     & \psnr & \ssim & \lpips & \psnr & \ssim & \lpips & \psnr & \ssim & \lpips & \psnr & \ssim & \lpips & \psnr & \ssim & \lpips & \psnr & \ssim & \lpips & \psnr & \ssim & \lpips \\
    \midrule
    RobustNeRF~\cite{sabour2023robustnerf}    & 17.54 & 0.496 & 0.383 & 15.65 & 0.318 & 0.576 & 23.04 & 0.764 & 0.244 & 20.39 & 0.718 & 0.251 & 20.65 & 0.625 & 0.391 & 20.54 & 0.578 & 0.366 & 19.64 & 0.583 & 0.369 \\
    NeRF On-the-go~\cite{ren2024nerf}           & 20.15 & 0.644 & 0.259 & 20.11 & 0.609 & 0.314 & 24.22 & 0.806 & 0.190 & 20.78 & 0.754 & 0.219 & 23.33 & 0.787 & 0.189 & 21.41 & 0.718 & 0.235 & 21.67 & 0.720 & 0.234 \\
    3DGS~\cite{kerbl3Dgaussians}                                     & 19.40 & 0.638 & 0.213 & 19.96 & 0.659 & 0.185 & 20.90 & 0.713 & 0.241 & 17.48 & 0.704 & 0.199 & 20.77 & 0.693 & 0.316 & 17.29 & 0.604 & 0.363 & 19.30 & 0.668 & 0.253 \\
    WildGaussian~\cite{kulhanek2024wildgaussians} & \third{20.43} & 0.653 & 0.255 & \third{20.81} & 0.662 & 0.215 & 24.16 & 0.822 & 0.139 & \third{21.44} & 0.800 & 0.138 & 23.82 & 0.816 & 0.138 & 22.23 & 0.725 & 0.206 & 22.16 & 0.746 & 0.182 \\
    DeSplat$^\ddagger$~\cite{wang2024desplat} & 19.59 & \third{0.715} & \second{0.175} & 20.27 & \third{0.685} & \third{0.175} & \first{26.05} & \first{0.885} & \second{0.095} & 20.89 & \third{0.815} & \third{0.115} & \second{26.07} & \first{0.905} & \second{0.095} & \third{22.59} & \first{0.845} & \second{0.125} & \third{22.58} & \third{0.813} & \second{0.130} \\
    Spotlesssplats~\cite{sabour2024spotlesssplats}  & \second{21.64} & \second{0.725} & \third{0.195} & \second{22.38} & \first{0.768} & \second{0.166} & \third{25.77} & \second{0.877} & \third{0.117} & \second{22.40} & \second{0.833} & \second{0.108} & \third{25.35} & \third{0.866} & \third{0.127} & \second{22.98} & \second{0.808} & \third{0.155} & \second{23.42} & \second{0.813} & \third{0.145} \\
    Ours & \first{22.31} & \first{0.746} & \first{0.163} & \first{22.40} & \second{0.764} & \first{0.139} & \second{25.94} & \third{0.869} & \first{0.078} & \first{22.88} & \first{0.850} & \first{0.087} & \first{26.59} & \second{0.886} & \first{0.089} & \first{23.35} & \third{0.799} & \first{0.124} & \first{23.91} & \first{0.819} & \first{0.113} \\
    \bottomrule

    \end{tabular}
    
    }

    \end{center}

\end{table*}

\input{figs/figurerobust}
\input{figs/neu3d}
\begin{table*}[htbp]
    \begin{center}
    \caption{Comparison dynamic modeling on Neu3D Dataset~\cite{li2022neural}. The \first{best}, \second{second best}, and \third{third best} are highlighted. Noticeably, our method shows a consistently better LPIPS score compared to baseline methods.}
    \label{table:neu3d}
    \scalebox{0.5}{
    \begin{tabular}{l
        ccc 
        ccc 
        ccc 
        ccc 
        ccc 
        ccc 
        | ccc 
        }
    \toprule
     & \multicolumn{3}{c}{Cut Beef} & \multicolumn{3}{c}{Cook Spinach} & \multicolumn{3}{c}{Sear Steak} & \multicolumn{3}{c}{Flame Steak} & \multicolumn{3}{c}{Flame Salmon} & \multicolumn{3}{c}{Coffee Martini} & \multicolumn{3}{|c}{Mean} \\
    \cmidrule(lr){2-4}\cmidrule(lr){5-7}\cmidrule(lr){8-10}\cmidrule(lr){11-13}\cmidrule(lr){14-16}\cmidrule(lr){17-19}\cmidrule(lr){20-22}
     & \psnr & \ssim & \lpips & \psnr & \ssim & \lpips & \psnr & \ssim & \lpips & \psnr & \ssim & \lpips & \psnr & \ssim & \lpips & \psnr & \ssim & \lpips & \psnr & \ssim & \lpips \\
    \midrule
    NeRFPlayer\cite{song2023nerfplayer} 
    & 31.83 & 0.928 & 0.119 
    & 32.06 & 0.930 & 0.116 
    & 32.31 & 0.940 & 0.111 
    & 27.36 & 0.867 & 0.215 
    & 26.14 & 0.849 & 0.233 
    & \first{32.05} & \third{0.938} & 0.111 
    & 30.29 & 0.909 & 0.151 \\
    HyperReel~\cite{attal2023hyperreel} 
    & \third{32.25} & 0.936 & \third{0.086} 
    & 31.77 & 0.932 & \third{0.090} 
    & 31.88 & 0.942 & \third{0.080} 
    & 31.48 & 0.939 & \third{0.083} 
    & 28.26 & \third{0.941} & 0.136 
    & 28.65 & 0.897 & 0.129 
    & 30.72 & 0.931 & 0.101 \\
    HexPlane~\cite{cao2023hexplane}    
    & 30.83 & 0.927 & 0.115 
    & 31.05 & 0.928 & 0.114 
    & 30.00 & 0.939 & 0.105 
    & 30.42 & 0.939 & 0.104 
    & 29.23 & 0.905 & \third{0.088} 
    & 28.45 & 0.891 & 0.149 
    & 30.00 & 0.922 & 0.113 \\
    KPlanes~\cite{fridovich2023k} 
    & 31.82 & \first{0.966} & 0.114 
    & \second{32.60} & \first{0.966} & 0.114 
    & \second{32.52} & \first{0.974} & 0.104 
    & \second{32.39} & \first{0.970} & 0.102 
    & \first{30.44} & \first{0.953} & 0.132 
    & \second{29.99} & \second{0.953} & 0.134 
    & \first{31.63} & \first{0.964} & 0.117 \\
    MixVoxels~\cite{wang2023mixed}
    & 31.30 & \second{0.965} & 0.111 
    & 31.65 & \second{0.965} & 0.113 
    & 31.43 & \second{0.971} & 0.103 
    & 31.21 & \second{0.970} & 0.108 
    & \second{29.92} & \second{0.945} & 0.163 
    & \third{29.36} & \third{0.946} & 0.147 
    & 30.81 & \second{0.960} & 0.124 \\
    SWinGS~\cite{shaw2023swings} 
    & 31.84 & 0.945 & 0.099 
    & 31.96 & 0.946 & 0.094 
    & 32.21 & 0.950 & 0.092 
    & \third{32.18} & 0.953 & 0.087 
    & \third{29.25} & 0.925 & 0.100 
    & 29.25 & 0.925 & \third{0.100} 
    & \third{31.12} & 0.941 & \third{0.095} \\
    4DGS~\cite{wu20244d} 
    & \first{32.66} & 0.946 & \second{0.053} 
    & \third{32.46} & 0.949 & \second{0.052} 
    & \third{32.49} & \third{0.957} & \second{0.041} 
    & \first{32.75} & 0.954 & \second{0.040} 
    & 29.00 & 0.912 & \second{0.081} 
    & 27.34 & 0.905 & \second{0.083} 
    & 31.12 & 0.937 & \second{0.058} \\
    Ours 
    & \second{32.56} & \third{0.957} & \first{0.042} 
    & \first{32.61} & \third{0.950} & \first{0.041} 
    & \first{33.20} & 0.956 & \first{0.035} 
    & \first{32.75} & \third{0.955} & \first{0.034} 
    & 29.23 & 0.916 & \first{0.068} 
    & 28.80 & 0.916 & \first{0.062} 
    & \second{31.52} & \third{0.942} & \first{0.047} \\
    \bottomrule
    \end{tabular}}
    \end{center}

\end{table*}

\input{figs/hypernerf}

\subsection{Implementation Details} \textbf{Initialization}  
 The scene boundary and the background gaussians are initialized from point clouds generated using COLMAP~\cite{schoenberger2016mvs,schoenberger2016mvs} or sensor perception~\cite{aria_data_tools} for Aria Sequences\cite{lv2024aria,pan2023aria,banerjee2024introducing}. The foreground Gaussians are initialized from randomly generated points distributed within this scene boundary.\\
\textbf{Coarse Training Stage:}  
During the coarse training stage, we disable the deformation module in the foreground branch and train both the foreground and background models for 1,000 iterations with short video clips and image collections or for iterations equal to sequence length for long captures.
\textbf{Fine Training Stage:}  
In the fine training stage, we jointly optimize the foreground and background branches end-to-end. For short video clips and image collections of less than 500 images, training iterations are set to 20k.
For input long video clips of a few thousand frames, the training iteration is set to 120k.
\subsection{Datasets}
\textbf{Egocentric video sequences} are with intensive camera wearer activities and varying illumination conditions, which pose challenges to scene modeling methods. We take one sequence from ADT~\cite{pan2023aria}, AEA~\cite{lv2024aria}, Hot3D~\cite{banerjee2024introducing}, and Epic-Field~\cite{tschernezki2024epic} dataset, respectively, ranging from 2800-5000 frames, to evaluate our method against baseline methods~\cite{kerbl3Dgaussians,sabour2024spotlesssplats,tschernezki2021neuraldiff} in diverse scenarios. For each sequence, every 1 out of 5 frames is held out during training.\\
\textbf{NerF On-the-Go Dataset}~\cite{ren2024nerf} comprises several hundred input images featuring moving distractors alongside a smaller set of clean images reserved for testing. We train our methods on the noisy occluded images and assess the quality of novel view synthesis on the clean hold-out set.\\
\textbf{Neu3D Dataset}~\cite{li2022neural} was captured using 15 to 20 static cameras recording relatively simple activities over 300 frames. Camera view 0 is the testing set, with the remaining views used for training.\\
\textbf{HyperNeRF Dataset}~\cite{park2021hypernerf} features real-world activities captured with smooth trajectories. However, as noted in~\cite{huang2023sc}, the camera poses are considerably inaccurate. Therefore, we focus primarily on qualitative visualizations for this dataset.
\subsection{Results}
To assess the performance of our method for the distractor-free scene reconstruction task in the presence of noisy inputs, we conduct evaluations on both egocentric videos and image collections. For egocentric video data~\cite{lv2024aria,banerjee2024introducing,pan2023aria,tschernezki2024epic}—which lack clean view references—we present qualitative comparisons with baseline methods~\cite{kerbl3Dgaussians,sabour2024spotlesssplats,tschernezki2021neuraldiff} in~\cref{fig:aria}. Compared to baseline methods~\cite{kerbl3Dgaussians,tschernezki2021neuraldiff,sabour2024spotlesssplats}, our method models high-quality distractor-free static background with accurate foreground separation. 
We additionally report video comparisons in our supplementary materials.
For image collections dataset Nerf-on-the-go\cite{ren2024nerf} with clean reference test views, we report detailed per-scene metrics including peak signal-to-noise ratio (PSNR), perceptual quality (LPIPS)~\cite{zhang2018unreasonable}, and structural similarity index (SSIM)~\cite{wang2004image} against baseline methods\cite{sabour2023robustnerf,ren2024nerf,kerbl3Dgaussians,kulhanek2024wildgaussians,sabour2024spotlesssplats,wang2024desplat} on the hold-out test set in~\cref{table:onthego}. Our methods generalize to image collections and achieve state-of-the-art results. Notably, our method consistently achieves significantly better LPIPS scores over the previous SOTA method SpotlessSplats~\cite{sabour2024spotlesssplats}. We show our method robustly handles occlusion and reconstructs fine static details compared to SpotlessSplats~\cite{sabour2024spotlesssplats}in~\cref{fig:qualionthego}. Additionally, our methods could naturally handle various input challenges, such as camera motion blur and lens flare, as shown in~\cref{fig:robust}.

Moreover, we compare our method's composed render quality with various baseline methods~\cite{cao2023hexplane,fridovich2023k,song2023nerfplayer,attal2023hyperreel,shaw2023swings,wang2023mixed,wu20244d} in~\cref{table:neu3d}, where our methods achieve consistently better LPIPS scores. We qualitatively show the dynamic reconstruction comparison and the rendering FPS of~\cite{wu20244d} and our method in~\cref{fig:qualidynerf}(on RTX4090), where our methods show better reconstructed fine details and better test-time rendering efficiency. Moreover, we compare our method with 4DGS~\cite{wu2022d} on HyperNeRF~\cite{park2021hypernerf} dataset in \cref{fig:HyperNerf}, showing that our method effectively regularizes gaussian movements with probabilistic controlled dynamic foreground representation and reduces unregularized moving artifacts. 

\section{Ablation study}
\label{sec:ablation}
\textbf{Brightness Control(BC)} is introduced to enhance the background branch's capacity to model non-Lambertian effects and mitigate dynamic-static ambiguities caused by varying illuminations, as shown in~\cref{fig:ablationaria}. w/o BC leads to downgraded performance in~\cref{table:ablationonthego}.\\
\textbf{Partial Opacity Reset(POR)} controls the gaussian density, facilitates floaters pruning, and mitigates local minima assignment. We show in~\cref{fig:ablationaria} and~\cref{table:ablationonthego} that this design leads to cleaner separation.\\
\textbf{Background Mask Element}($m_b^{\prime}$) is introduced to promote cleaner separation and discourage mid-range probabilities. Though the improvements are not significant for image collections with good initializations, it leads to better dynamic-static modeling and separation results as shown in~\cref{fig:ablationaria}. \\
\textbf{Loss} $\mathcal{L}_\text{depth}$ is introduced to promote reconstruction with smooth background geometry and loosely regularize foreground and background depth prediction. As shown in~\cref{fig:ablation_neu3d}, this component efficiently prevents unconstrained gaussians from occluding test-time render for sparse, fixed camera input. $\mathcal{L}_\text{depth}$ also leads to better rendering quality as shown in~\cref{table:ablationonthego}.

\input{figs/ablation_aria}
\input{figs/ablation_Neu3d}
\begin{table}[t]
    \begin{center}
    \caption{Ablation study on Nerf-on-the-go dataset\cite{ren2024nerf}}
    \label{table:ablationonthego}
    \scalebox{0.7}{
    \begin{tabular}{cccc}
    \toprule 
    Sequence from & \psnr & \ssim & \lpips \\
    \midrule

    w.o BC    & 23.54 & 0.814 & 0.118  \\
    w.o POR  &23.56 &0.814 &  0.117\\
    w.o $\mathcal{L}_\text{depth}$      & 23.68 & 0.816 & 0.113 \\
    w.o. $m_b^\prime$ & 23.83 & 0.817 & 0.115\\
    Ours & \textbf{23.91} & \textbf{0.819} & \textbf{0.113} \\ 
    \bottomrule
    \end{tabular}}
    \end{center}
\end{table}

\section{Conclusion}
\label{sec:conclusion}
This paper proposes DeGauss to robust decompose dynamic-static elements in the scene with gaussian splatting. With decoupled dynamic-static gaussian branches controlled by mask attributes rasterized by foreground gaussians, our method achieves flexible yet accurate dynamic-static decomposition that widely generalizes to various scenarios, leading to clean distractor-free static scene modeling and high-quality and efficient dynamic scene modeling.
 
\noindent\textbf{Acknowledgements.} 
 The authors would like to sincerely thank Siwei Zhang for the rigorous proof-reading and Hui Zhang for the discussion.
{
    \small
    \bibliographystyle{ieeenat_fullname}
    \bibliography{main}
}
\clearpage
\maketitlesupplementary
\appendix
\section{Detailed Loss Function Formulation}
\label{sec:appendix:loss}
Loss function design is important to maintain the balance of the dynamic-static decomposition task. For example, directly adding SSIM loss could improve the overall reconstruction quality but often leads to a larger gradient magnitude in the static region with fine details. As a result, this often leads to the over-expressiveness of foreground gaussians that undesirably models the static fine details. As the densification of process of gaussian is controlled by loss gradient magnitude, we propose a loss function that comprises two components $\mathcal{L}_\text{main}$ and $\mathcal{L}_\text{uti}$to decouple parameter updates and the adaptive densification process. While both $\mathcal{L}_\text{main}$ and $\mathcal{L}_\text{uti}$ contribute to the background and foreground gaussian feature updates, only the gradient of $\mathcal{L}_\text{main}$ is used for the densification process. The main loss component is defined as:
\begin{equation}
    \mathcal{L}_\mathrm{main} = \mathcal{L}_1 + \mathcal{L}_\mathrm{reg} + \mathcal{L}_\text{diversity} + \mathcal{L}_{f} + \mathcal{L}_{b} + \mathcal{L}_\mathrm{depth},
    \label{equ:mainloss}
\end{equation}
where 
\[
\mathcal{L}_\text{1} = \lVert \hat{\mathbf{C}} - \mathbf{C}_{gt} \rVert_1
\]
denotes the \(\mathcal{L}_1\) loss between the fully composed rendered image \(\hat{\mathbf{C}}\) and the ground truth image \(\mathbf{C}_{gt}\). The regularization loss \(\mathcal{L}_\mathrm{reg}\) enforces time smoothness and k-plane total variations, following the settings in \cite{cao2023hexplane, TiNeuVox, fridovich2023k, wu20244d}.
Furthermore, to encourage a higher foreground probability \(\mathbf{P}_f\) for the foreground render \(\hat{\mathbf{C}}_f\) at region which exhibits significant structural differences relative to the detached background render \(\bar{\mathbf{C}}_b\), similar to \cite{ren2024nerf}, we employ a diversity loss based on the structural component of the SSIM loss :
\begin{equation}
	\mathcal{L}_\text{diversity}(\mathbf{C}_f, \bar{\mathbf{C}}_b) = \mathbbm{1}_{\{\mathbf{P}_f > \mathbf{P}_\tau\}} \cdot \frac{\sigma_{\mathbf{C}_f \bar{\mathbf{C}}_b} + c_3}{\sigma_{\hat{\mathbf{C}}_f}\,\sigma_{\bar{\mathbf{C}}_b} + c_3},
\end{equation}
where \(\mathbbm{1}_{\{\mathbf{P}_f > \mathbf{P}_\tau\}}\) is the indicator function and \(\mathbf{P}_\tau\) is the probability threshold, \(\sigma\) denotes the variance, and \(c_3\) is a constant to stabilize the loss. To refine the regions assigned to the background and foreground, we further introduce updating losses \(\mathcal{L}_f\) and \(\mathcal{L}_b\), defined as:

\begin{align}
\mathcal{L}_{e} &= \mathbbm{1}_{\{\mathbf{P}_e > \mathbf{P}_\tau\}} \Bigl( \lVert \hat{\mathbf{C}}_{e} - \mathbf{C}_{gt} \rVert_1\nonumber \\
&+ 0.1 \mathcal{L}_\text{SSIM}(\hat{\mathbf{C}}_{e}, \mathbf{C}_{gt}) \Bigr),
\quad e \in \{f,b\}.
\end{align}

This loss term is scaled down 4 times compared to the $\mathcal{L}_1$ loss between foreground render and background render to suppress their contribution to gaussian densification process.
Additionally, to softly regularize the spatial relationship between foreground-background  gaussians and encourage a distractor-free background reconstruction, we introduce depth-related loss $\mathcal{L}_\text{depth}$, defined as:
\begin{equation}
    \mathcal{L}_\mathrm{depth} = \mathcal{L}_\mathrm{smooth} + \mathcal{L}_\mathrm{sep},
\end{equation}
where \(\mathcal{L}_\mathrm{smooth}\) is an edge-aware total variation loss \cite{turkulainen2024dnsplatter, heise2013pm} that encourages smooth depth predictions for static background Gaussians, particularly in regions with small color variance:
\begin{equation}
\begin{aligned}
\mathcal{L}_\mathrm{smooth} &= \frac{1}{N} \sum_{i,j} \Bigl( 
    \bigl|D_{b_{ij}} - D_{b_{ij+1}}\bigr| \cdot e^{-\|\mathbf{C}_{gt_{ij}} - \mathbf{C}_{gt_{ij+1}}\|_1} \\
    &\quad + \bigl|D_{b_{ij}} - D_{b_{i+1j}}\bigr| \cdot e^{-\|\mathbf{C}_{gt_{ij}} - \mathbf{C}_{gt_{i+1j}}\|_1} 
\Bigr).
\end{aligned}
\end{equation}
Here, \(N\) denotes the total number of pixels, and \(D_{b_{ij}}\) represents the depth value at pixel \((i,j)\) of the rendered background depth image, normalized by the scene bounding box to account for the scale ambiguity of colmap\cite{schoenberger2016sfm} reconstruction. Moreover, the depth separation loss is defined as:
\begin{align}
    \mathcal{L}_\mathrm{sep} = & \mathbbm{1}_{\{\mathbf{P}_f > \mathbf{P}_\tau\}} \Bigl(\sum_{i,j} \max(D_{f_{ij}} - D_{b_{ij}}, 0) \Bigr) \\ &+ \mathbbm{1}_{\{\mathbf{P}_b > \mathbf{P}_\tau\}} \Bigl(\sum_{i,j} \max(D_{f_{ij}} - D_{b_{ij}}, 0) \Bigr).
\end{align}
The first loss term encourages the rendered foreground to be positioned closer to the camera, thereby preserving occlusion relationships with the static background. In addition, the second loss term pushes the utility gaussians with low foreground render contributions to be further away from the camera to prevent their presence during novel view rendering. This term efficiently regularizes floaters for datasets with sparse fixed camera input as Neu3D dataset~\cite{li2022neural}.

 \(\mathcal{L}_\mathrm{uti}\) is introduced to stabilize training, promote fine reconstruction, and enhance separation without contributing to the densification process:
\begin{equation}
      \mathcal{L}_\mathrm{ulti} = \mathcal{L}_\mathrm{SSIM}(\hat{\mathbf{C}}, \mathbf{C}_{gt}) + \mathcal{L}_\mathrm{entropy} + \mathcal{L}_\mathrm{brightness} + \mathcal{L}_s.
\end{equation}
The SSIM loss \(\mathcal{L}_\mathrm{SSIM}\), computed between the composed render \(\hat{\mathbf{C}}\) and the ground truth image \(\mathbf{C}_{gt}\), improves reconstruction quality of fine detailed region; Additionally, the entropy loss is defined as a binary cross-entropy loss that encourages the foreground probability \(\mathbf{P}_f\) to converge toward either 0 or 1:
\begin{equation}
     \mathcal{L}_\mathrm{entropy} = -\sum_{N} \, \mathbf{P}_f \cdot \log(\mathbf{P}_f).
\end{equation}
Furthermore, to promote the update of brightness control mask $\hat{\mathbf{B}}$ in the early stage, we define the brightness loss as:
\begin{equation}
	\mathcal{L}_\mathrm{brightness} = \alpha \cdot \lVert \hat{B} \ast \bar{\mathbf{C}}_b - \mathbf{C}_{gt} \rVert_1 + (1 - \alpha) \cdot \lVert \hat{B} - 1 \rVert_1,
\end{equation}
where \(\bar{\mathbf{C}}_b\) denotes the novel view rendered from the background branch (detached from gradient propagation), and \(\alpha\) is a coefficient that increases linearly with training iterations. The first term ensures an accurate prediction of the brightness control mask, while the second term acts as a regularizer. Finally, the scale loss \(\mathcal{L}_s\) penalizes spiky Gaussians, as defined in \cite{xie2023physgaussian}.

The loss coefficients set to balance each loss term is set to 4 for main $\mathcal{L}_1$ loss, 1 for $\mathcal{L}_f$ and $\mathcal{L}_b$, 0.01 for $\mathcal{L}_\text{entropy}$ and 0.1 for the rest components.

\begin{figure*}[t]
	\centering
	\includegraphics[width=1\linewidth]{images/supplementary_neu3d.pdf}
	\caption{Visualization for Flame Steak Sequence of Neu3D~\cite{li2022neural} dataset. Our method achieves accurate dynamic-static decomposition with high reconstruction quality.}
	\label{fig:morequaliNeu3D}
\end{figure*}
\begin{figure*}[h]
	\centering
	\includegraphics[width=1\linewidth]{images/hypernerf_extra.pdf}
	\caption{Visualization of dynamic modeling on peel banana and chicken sequence on HyperNerf Vrig dataset~\cite{park2021hypernerf} dataset. Our methods reconstruct high-quality dynamic scenes with an efficient dynamic-static decoupled representation.}
	\label{fig:morequalihyperNerf}
\end{figure*}

\begin{figure}[t]
	\centering
	\includegraphics[width=1\linewidth]{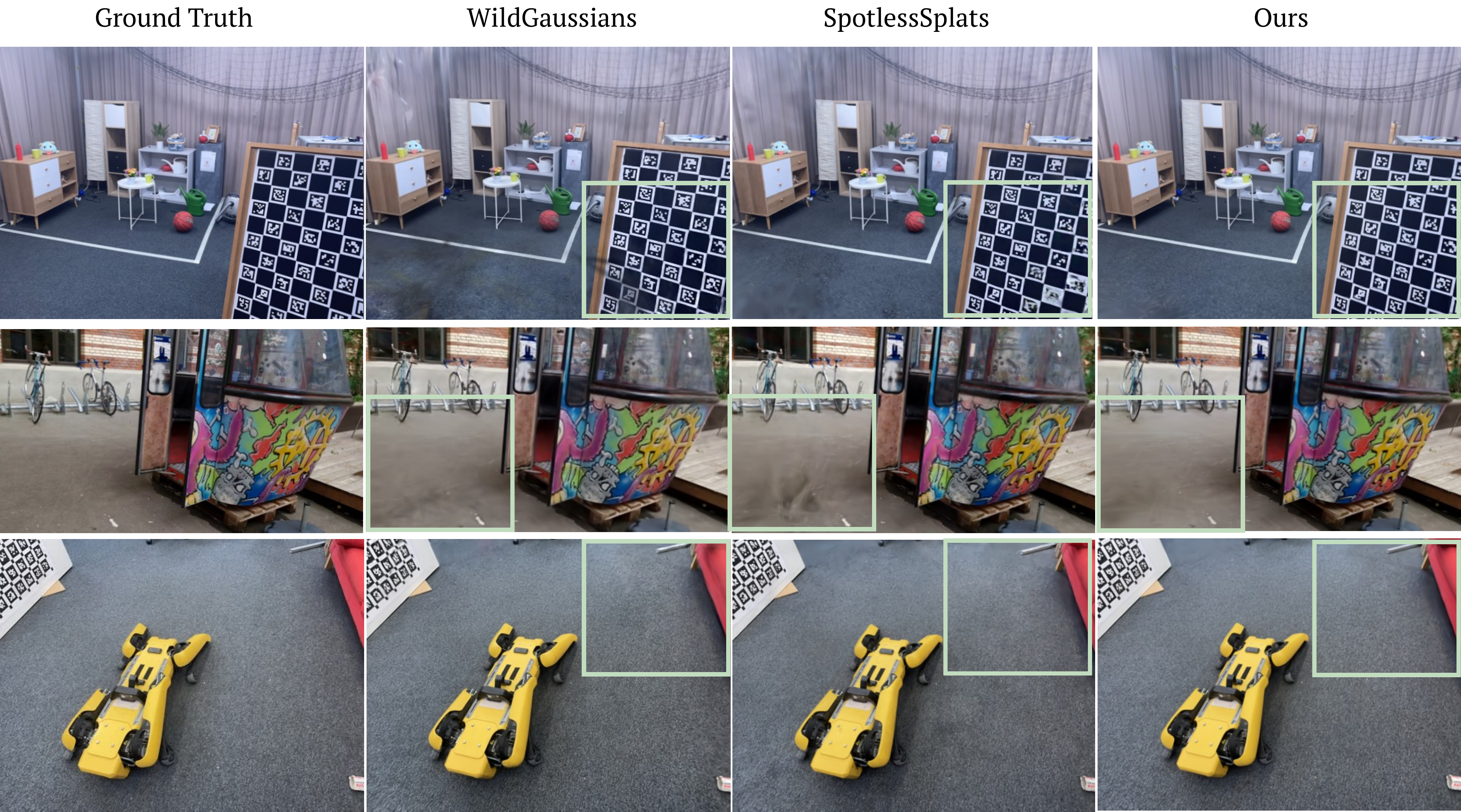}
	\caption{Qualitative comparison of baseline methods\cite{kulhanek2024wildgaussians,sabour2024spotlesssplats} on Nerf-On-the-go dataset.}
	\label{fig:qualicompareonthego}
\end{figure}

\section{Additional Pruning for Dynamic Scene modeling}
In our setting, there are utility gaussians that do not contribute to dynamic rendering but are utilized for probabilistic mask and brightness control mask rasterization. Therefore, we could optionally further control the number of utility gaussians with foreground visibility-based pruning.

Specifically, a Gaussian is discarded if the maximum value of the product of its opacity $\sigma$ and its foreground mask elements $m_f^\prime$—computed across all input views and timestamps—falls below a predefined threshold \(\tau\). This procedure effectively eliminates Gaussians that contribute negligibly to the overall dynamic representation for dynamic scene modeling tasks.
\section{Detailed Dataset preparation}
\textbf{Aria Glass Recordings}~\cite{pan2023aria,lv2024aria, banerjee2024introducing} feature egocentric video captured at 20-30 FPS, encompassing intensive human-object, human-scene, and human-human interactions, along with challenges such as rapid camera motion and motion blur. We used NerfStudio~\cite{nerfstudio} to preprocess fisheye camera frames from Project Aria, with camera mask and distortion parameters, and camera vignetting mask provided~\cite{aria_data_tools,gu2025egolifter}. The original resolution of an Aria frame is \(1415 \times 1415\) after fisheye undistortion. To balance rendering quality and speed and avoid excessive training time for the nerf baseline neuraldiff, we downsample the frames to $707\times707$. The first 50 frames of each sequence are omitted to allow the camera stream to stabilize. 

\textbf{Epic-field Dataset} \cite{tschernezki2024epic} builds upon the EPIC-Kitchen dataset \cite{damen2020epic}, which comprises long egocentric video recordings of human activities in a kitchen recorded at 50 FPS. We use the point clouds and camera poses provided in \cite{tschernezki2024epic}. To keep a consistent frame rate with aria recordings, we
take testing segments of 10,000 consecutive frames and downsample by 2, which leads to 5000 frames in the end.

\textbf{NerF On-the-Go Dataset}~\cite{ren2024nerf} we prepare the Nerf On-the-go dataset following the setting of SpotlessSplats~\cite{sabour2024spotlesssplats}. The dataset was originally captured with high-resolution images and downscaled 4 times for patio set and 8 times for others, following~\cite{ren2024nerf,kulhanek2024wildgaussians,sabour2024spotlesssplats}. We follow the camera undistortion setting of~\cite{sabour2024spotlesssplats}.
\textbf{Neu3D Dataset}~\cite{li2022neural}  Following the setup of \cite{wu20244d}, the resolution is downscaled to $1352\times1014$.
We compute the camera poses and generate a dense point cloud using COLMAP~\cite{schoenberger2016sfm,schoenberger2016mvs} based on the first frame of each video.\\
\textbf{HyperNerf dataset} As noted in~\cite{huang2023sc}, the camera poses are considerably inaccurate, which diminishes the reliability of quantitative comparisons. Therefore, we run colmap\cite{schoenberger2016sfm} to recompute camera poses and focus primarily on qualitative visualizations for this dataset.
\\
\section{Implementation Details}
\textbf{Initialization} During initialization, for the background branch, Gaussians are derived from point clouds generated using COLMAP \cite{schoenberger2016mvs,schoenberger2016mvs} or from sparse perception point clouds provided by the ARIA project \cite{aria_data_tools}. The scene boundary is determined based on the range of the background points, with an additional padding equal to 0.3 times the diagonal length of the camera trajectory. Foreground Gaussians are initialized from randomly generated points within this 3D scene boundary.

\textbf{Coarse Training Stage} In the coarse training stage, we disable the deformation module in the foreground branch and train both the foreground and background models for 1,000 iterations. For longer sequences (containing thousands of frames), the number of coarse training iterations is adjusted so that each image is processed exactly once. During this stage, the standard color loss \(\mathcal{L}_\mathrm{1}\) in Equation~\ref{equ:mainloss} is replaced by a combination of foreground and background losses:
\[
\mathcal{L}_\mathrm{coarse} = \lVert (\mathbf{P}_f \ast \mathbf{C}_f + \mathbf{P}_b \ast \hat{B} \ast \bar{\mathbf{C}}_b) - 0.9\, \mathbf{C}_{gt} \rVert_1 + \lVert \mathbf{C}_b - \mathbf{C}_{gt} \rVert_1.
\]
The discount factor of 0.9 applied to the ground truth further regularizes the expressiveness of the foreground Gaussians, particularly in featureless regions (e.g., walls) that are often associated with poor structural reconstruction in COLMAP.

\textbf{Fine Training Stage} In the fine training stage, we jointly optimize the foreground and background branches. For short video clips and image collections, training is performed for 20,000 iterations; for longer video clips, training extends to 120,000 iterations.\\
\textbf{Parameters set up} We generally follow the parameter set up in~\cite{wu20244d}. With the basic resolution of Hexplane set to 256 for egocentric recordings and 64 for other scenes, upsampled by 2 and 4 The learning rate of set to Hexplane is set to $6 \times 10^{-4}$ and decays to $2 \times 10^{-5}$ during training.  The deformation learning rate is set to $1.6\times10^{-4}$ and decays to $1.6\times10^{-5}$ during training. The deformation learning rate for mask update is set to $1.6\times10^{-5}$ and decays to $1.6\times10^{-6}$ during training. Generally, the batch size is set to 2 as~\cite{wu20244d}. For low-resolution image collections in~\cite{ren2024nerf}, we set the batch size to 4 for the dynamic branch and additionally accumulate the update of 4 batches for the background gaussians to account for the low resolution and loose temporal correlations. \\
\textbf{Baseline Evaluation} For 4DGS~\cite{wu20244d} experiments, we follow the instruction of their official repo and dataset preparation. For the HyperNerf~\cite{park2021hypernerf} dataset, we use the colmap calculated camera poses and point cloud for initialization, the same as our method. For experiments with 3DGS~\cite{yang2023deformable3dgs} and~\cite{sabour2024spotlesssplats}, we use the official repo of~\cite{sabour2024spotlesssplats} and follow their setup. For the Nerf on-the-go dataset~\cite{ren2024nerf}, EPIC-Field dataset, we use standard colmap initialization. For Aria sequences, the sensor perception point clouds are without color, which leads to unstable initialization for ~\cite{sabour2024spotlesssplats}. Therefore, we triangulate with COLMAP~\cite{schoenberger2016sfm} using camera poses provided by~\cite{aria_data_tools} to obtain colored point cloud to better evaluate this method.
\section{Efficiency Analysis on Neu3D~\cite{li2022neural} dataset}
Our dynamic-static hybrid representation enables:\textbf{(a) Much higher FPS}:  The time critical process of deformation prediction in 4DGS scales with the number of dynamic Gaussians. Table~\ref{table:reneu3d} shows we render \textbf{3× faster} than 4DGS with superior quality by minimal dynamic element modeling with our dynamic-static decoupling design. \textbf{(b) Better Quality} We achieve much higher LPIPS and finer details in static (no stray motion) and dynamic (better handling disappearing gaussians), as reported in our paper and project page: \url{https://batfacewayne.github.io/DeGauss.io/}. Even on Coffee Martini, Flame Salmon with very far objects that poses challenges to gaussian splatting methods, our LPIPS and details remain best. \textbf{(c) Applications} The decoupled static with 3DGS seamlessly enables diverse applications as editing/styling.
\begin{table}[t]
  \begin{center}
    \footnotesize
    \setlength\tabcolsep{1.2pt}
    \caption{Quality and efficiency evaluation on all scenes of Neu3D~\cite{li2022neural} dataset tested on a RTX4090.$\dagger$: trained densify grad threshold $\times2$ to reduce number of gaussians.}
    \label{table:reneu3d}
       \scalebox{0.7}{
    \begin{tabular}{c|ccc|c|cc}
      \toprule 
      Method & {\footnotesize{PSNR($\uparrow$)}} & {\footnotesize{SSIM($\uparrow$)}} & {\footnotesize{LPIPS($\downarrow$)}} & {\footnotesize{Training Time($\downarrow$)}} & {\footnotesize{FPS($\uparrow$)}} & {\footnotesize{Dyna. Gaussian num($\downarrow$)}} \\
      \midrule
      NeRFPlayer~\cite{song2023nerfplayer} &30.29 & 0.909 & 0.151 &6 hours&0.045&-\\
      HyperReel~\cite{attal2023hyperreel} & 30.72 & 0.931 & 0.101 &9 hours&2.0&-\\
      HexPlane~\cite{cao2023hexplane}& 30.00 & 0.922 & 0.113 & 12 hours &0.2 & -\\
      KPlanes~\cite{fridovich2023k}& \first{31.63} & \first{0.964} & 0.117& 5.0 hours&0.3&-\\
      \midrule
    SWinGS~\cite{shaw2023swings}& 31.12 & \third{0.941} & 0.095& - & \second{71} &-\\
      4DGS~\cite{wu20244d}& 31.12 &0.937 &\second{0.058} & \third{0.85 hours}& 53 & 124,197\\
    4DGS$^{\dagger}$~\cite{wu20244d}&28.72 &0.919 &\third{0.078} & \second{0.67 hours}& \third{68} & \third{62298}\\
    \midrule
    Ours & \third{31.52} & \second{0.942} & \first{0.047} & {2.1 hours}& \second{71} & \second{56,533}\\
  Ours$^{\dagger}$& \second{31.56} & \second{0.942} & \first{0.047} & {2 hours} & \first{157} & \first{22,479}\\
      \bottomrule
    \end{tabular}
    }
  \end{center}
\end{table}
\section{Strict Monocular Input}
\begin{figure}[h]
  \centering
  \includegraphics[width=1\linewidth]{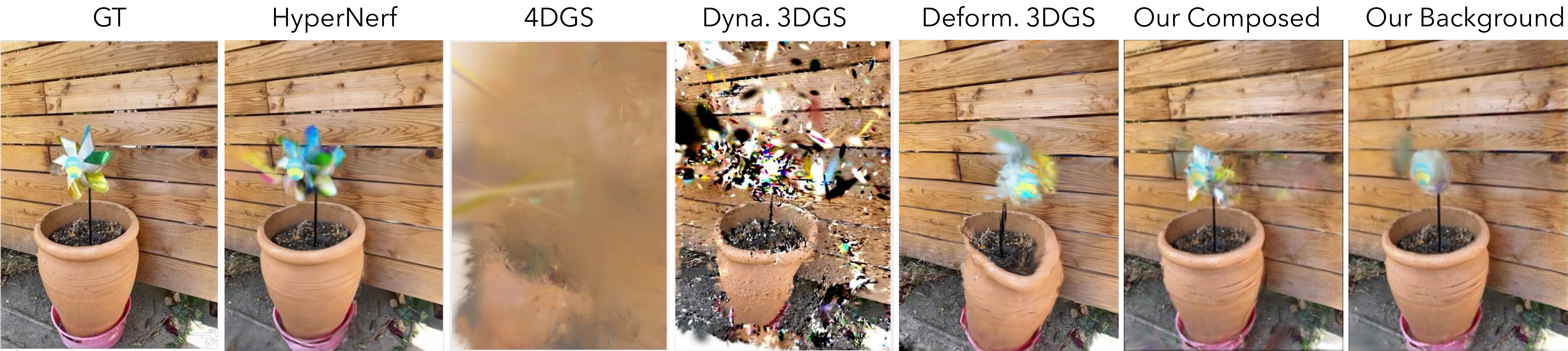}
   \caption{Comparison with baseline methods on novel view synthesis with causal strict monocular input of dycheck-iphone dataset~\cite{gao2022monocular}.}
   \label{fig:redycheck}
\end{figure}

Monocular reconstruction is extremely challenging. And compared to NeRF methods~\cite{park2021hypernerf}, dynamic gaussian methods~\cite{luiten2023dynamic,wu20244d,yang2023deformable3dgs} are highly expressive but much harder to regularize, generalizing poorly to novel views~\cite{stearns2024dynamic}(~\cref{fig:redycheck}). This actually enables our gaussian-based \textit{decoupled design}, to fast and robustly separate dynamic/static modeling for a wide range of inputs, and our explicit static modeling leads to much better generalizability of novel view synthesis for dynamic scene modeling(~\cref{fig:redycheck}).
\section{Additional Experiment on Bonn RGBD dataset~\cite{palazzolo2019iros}}
\begin{figure}[h]
  \centering
  \includegraphics[width=1\linewidth]{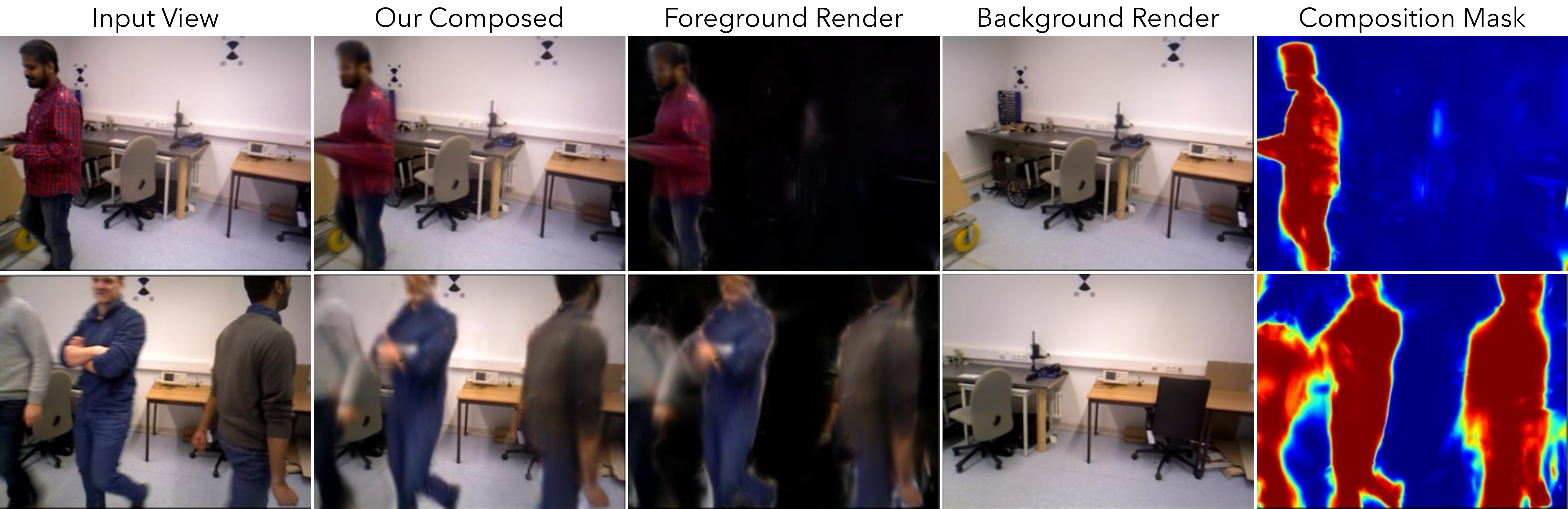}
   \caption{Evaluation on 928 frames long highly dynamic \textit{Crowd} scene of Bonn RGBD dataset~\cite{palazzolo2019iros}(with only RGB as input).}
   \label{fig:rebonn}
\end{figure}

To further demonstrate generalizability our method, we evaluate on the \textit{Crowd} scene of a SLAM dataset-Bonn RGBD~\cite{palazzolo2019iros}, preprocessed with SFM and MVS pipeline of~\cite{schoenberger2016sfm,schoenberger2016mvs}. We qualitatively show the distractor-free static scene reconstruction and dynamic-static decoupling results in~\cref{fig:rebonn}. 
\\

\section{Additional Experiment on RobustNerf~\cite{sabour2023robustnerf} dataset}
\begin{figure}[h]
  \centering
  \includegraphics[width=1\linewidth]{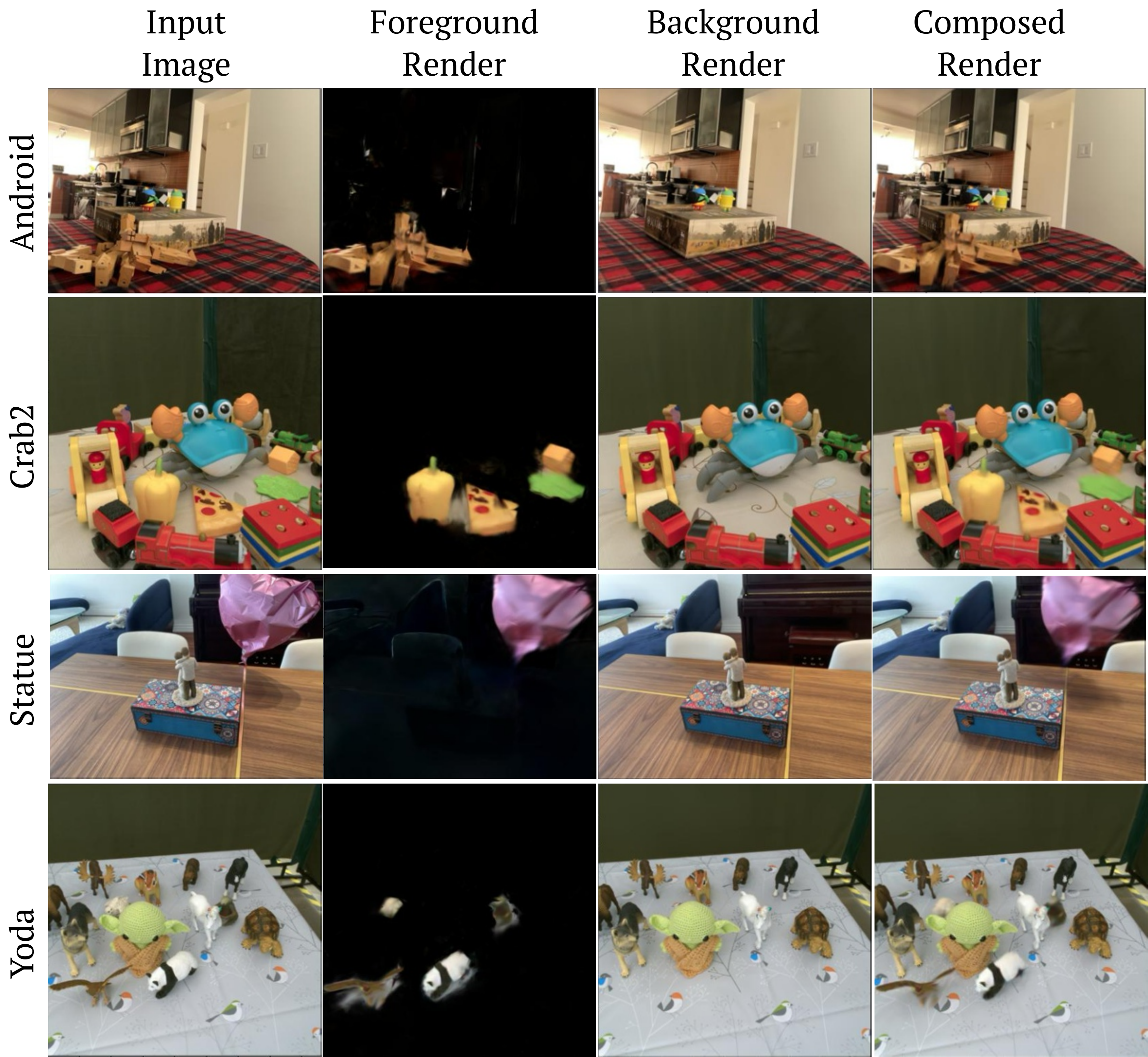}
   \caption{Qualitative result on RobustNerf dataset~\cite{sabour2023robustnerf}.}
   \label{fig:rerobust}
\end{figure}

We additionally report the performance of our method on RobustNerf~\cite{sabour2023robustnerf} in~\cref{table:rerobustnerf} and~\cref{fig:rerobust}.

\begin{table}[t]
  \begin{center}
    \footnotesize
    \setlength\tabcolsep{2pt}
    \caption{Quantitative results on RobustNerf~\cite{sabour2023robustnerf} dataset. Our method shows best overall performance and significantly better LPIPS score over all baseline methods.}
    \label{table:rerobustnerf}
    \scalebox{0.6}{
    \begin{tabular}{l|ccc|ccc|ccc|ccc}
      \toprule
      Method & \multicolumn{3}{c|}{Android} & \multicolumn{3}{c|}{Crab2} & \multicolumn{3}{c|}{Statue} & \multicolumn{3}{c}{Yoda} \\
      & PSNR$\uparrow$ & SSIM$\uparrow$ & LPIPS$\downarrow$
      & PSNR$\uparrow$ & SSIM$\uparrow$ & LPIPS$\downarrow$
      & PSNR$\uparrow$ & SSIM$\uparrow$ & LPIPS$\downarrow$
      & PSNR$\uparrow$ & SSIM$\uparrow$ & LPIPS$\downarrow$\\
      \midrule
      3DGS~\cite{kerbl3Dgaussians} & 23.32 & 0.794 & \third{0.159} & \third{31.76} & \third{0.925} & \third{0.172} & 20.83 & \third{0.830} & \third{0.148} & 28.92 & \third{0.905} & \third{0.192} \\
      WildGaussians~\cite{kulhanek2024wildgaussians} & \first{24.67} & \first{0.828} & \second{0.151} & 30.52 & 0.909 & 0.213 & \second{22.54} & \first{0.863} & \second{0.129} & \third{30.55} & \third{0.905} & 0.202  \\SpotLessSplat~\cite{sabour2024spotlesssplats} & \third{24.20} & \third{0.810} & \third{0.159} & \second{33.90} & \second{0.933} & \second{0.169} & \third{21.97} & 0.821 & 0.163 & \first{34.24} & \second{0.938} & \second{0.156}  \\
      Ours & \second{24.54} & \second{0.813} & \first{0.083} & \first{34.48} & \first{0.952} & \first{0.076} & \first{23.08} & \second{0.861} & \first{0.097} & \second{33.48} & \first{0.947} & \first{0.082}\\
      \bottomrule
    \end{tabular}
    }
  \end{center}
\end{table}

\section{Additional Visualizations}
We show additional visualizations in~\cref{fig:morequaliNeu3D}, ~\cref{fig:morequalihyperNerf} and~\cref{fig:qualicompareonthego}.
\section{Discussion}
\paragraph{Dynamic-Static Elements.}
While our method effectively handles semi-static objects, there is an inherent ambiguity when certain subjects—like people or objects—remain static most of the time in long video recordings.
In this work, we focus on a self-supervised approach that ensures robust decomposition across diverse scenarios.
For specific downstream applications, it may be beneficial to integrate our method with additional semantic information for even more accurate separation.

\paragraph{Camera Pose Optimization.}
Our approach generally assumes reasonably accurate camera poses to facilitate static-dynamic decomposition.
Nonetheless, we observe that even when camera poses are suboptimal (as in HyperNeRF~\cite{park2021hypernerf}), our method can still separate dynamic and static regions.
An interesting direction for future research is to leverage our predicted masks to optimize camera poses based on regions identified as static.

\paragraph{Efficient Dynamic Scene Representation.}
In this work, we showed that we could achieve high-quality and efficient dynamic representation by a decoupled dynamic-static gaussian representation, which largely reduces the number of gaussian in the time-consuming deformation step. However, as there are numerous utility gaussian to model probabilistic and brightness control mask.
Exploring ways to minimize this overhead could be a promising avenue for future work.

\end{document}